\documentclass[conference]{IEEEtran}
\IEEEoverridecommandlockouts
\usepackage{cite}
\usepackage{subcaption}
\usepackage{array,multirow}
\usepackage{subfig}
\usepackage{amsmath,amssymb,amsfonts}
\usepackage{algorithmic}
\usepackage{graphicx}
\usepackage{textcomp}
\usepackage{xcolor}
\usepackage{multirow}
\def\BibTeX{{\rm B\kern-.05em{\sc i\kern-.025em b}\kern-.08em
    T\kern-.1667em\lower.7ex\hbox{E}\kern-.125emX}}
\usepackage{listings}
\lstdefinelanguage[RISC-V]{Assembler}
{
  alsoletter={.}, 
  alsodigit={0x}, 
  morekeywords=[1]{ 
    lb, lh, lw, lbu, lhu,
    sb, sh, sw,
    sll, slli, srl, srli, sra, srai,
    add, addi, sub, lui, auipc,
    xor, xori, or, ori, and, andi,
    slt, slti, sltu, sltiu,
    beq, bne, blt, bge, bltu, bgeu,
    j, jr, jal, jalr, ret,
    scall, break, nop
  },
  morekeywords=[2]{ 
    .align, .ascii, .asciiz, .byte, .data, .double, .extern,
    .float, .globl, .half, .kdata, .ktext, .set, .space, .text, .word
  },
  morekeywords=[3]{ 
    zero, ra, sp, gp, tp, s0, fp,
    t0, t1, t2, t3, t4, t5, t6,
    s1, s2, s3, s4, s5, s6, s7, s8, s9, s10, s11,
    a0, a1, a2, a3, a4, a5, a6, a7,
    ft0, ft1, ft2, ft3, ft4, ft5, ft6, ft7,
    fs0, fs1, fs2, fs3, fs4, fs5, fs6, fs7, fs8, fs9, fs10, fs11,
    fa0, fa1, fa2, fa3, fa4, fa5, fa6, fa7
  },
  morecomment=[l]{;},   
  morecomment=[l]{\#},  
  morestring=[b]",      
  morestring=[b]'       
}

\definecolor{mauve}{rgb}{0.58,0,0.82}

\begin{document}

\title{IzhiRISC-V - a RISC-V-based Processor with Custom ISA Extension for Spiking Neuron Networks Processing with Izhikevich Neurons\\
}

\author{\IEEEauthorblockN{1\textsuperscript{st} Wiktor J. Szczerek}
\IEEEauthorblockA{\textit{Department of Computer Science} \\
\textit{KTH Royal Institute of Technology}\\
Stockholm, Sweden \\
ORCID: 0009-0005-5561-440X}
\and
\IEEEauthorblockN{2\textsuperscript{nd} Artur Podobas}
\IEEEauthorblockA{\textit{Department of Computer Science} \\
\textit{KTH Royal Institute of Technology}\\
Stockholm, Sweden \\
ORCID: 0000-0001-5452-6794}
}

\maketitle

\begin{abstract}
Spiking Neural Network processing promises to provide high energy efficiency due to the sparsity of the spiking events. However, when realized on general-purpose hardware -- such as a RISC-V processor -- this promise can be undermined and overshadowed by the inefficient code, stemming from repeated usage of basic instructions for updating all the neurons in the network. One of the possible solutions to this issue is the introduction of a custom ISA extension with neuromorphic instructions for spiking neuron updating, and realizing those instructions in bespoke hardware expansion to the existing ALU. In this paper, we present the first step towards realizing a large-scale system based on the RISC-V-compliant processor called IzhiRISC-V, supporting the custom neuromorphic ISA extension.
\end{abstract}


\section{Introduction}
\label{sec:intro}

With Dennard’s scaling ending~\cite{bohr200930} and Moore’s law~\cite{theis2017end} coming to an (expected) halt, how to continue providing high-performant yet power-efficient computation in the decades to come is today an open and unsolved research question. Out of the many proposed post-Moore technologies~\cite{shalf2020future}, such as quantum or adiabatic computing, perhaps the most salient options are those that are inspired by the human brain—neuromorphic computing~\cite{schuman2017survey,szczerek2025quarter}.
A neuromorphic computer is a computer system that is inspired by the animal brain, and is (to a large extent) composed of programmable neurons, synapses, and axons (delays). Most of the time (albeit not always), communication within a neuromorphic system is performed using sparse events called \textit{spikes} (or post-synaptic potentials, PSPs), and the neuromorphic system executes a spiking neural network~\cite{maass1997networks} (SNN) in silico. By mapping an application onto such an SNN, problems can sometimes be solved faster or more energy-efficiently than using a classical, imperative, von Neumann-based computer. Inspiring examples include LASSO, graph search, or constraint satisfaction problems, which have been shown to be more energy efficient and reach an order of magnitude faster time to solution compared to CPUs~\cite{davies2021advancing}.

In this paper, we explore how classical computer systems can be augmented with the power to perform neural and synoptical updates at high performance with a low silicon footprint. We design IzhiRISC-V -- an Application-Specific Instruction Set Processor (ASIP) that is based around a 3-stage RISC-V pipelined processor (RV32IMZ extensions), which we enhanced by integrating a custom neural processing unit (NPU) and a neuron decay unit (DCU). Our NPU targets fixed-precision single-cycle Euler updates of the Izhikevich model~\cite{izhikevich2003simple}, while the DCU is capable of AMPA-receptor decay of input current. 

We claim the following contributions:

\begin{itemize}
\item  We propose a set of instruction set architecture (ISA) extensions to RISC-V, allowing custom neural instructions to be included,
\item  We design a prototype architecture that includes said ISA neural extensions, showing how we can have single-cycle Izhikevich ODE integration and AMPA-like current dynamics,
\item  We quantify the resource, frequency, energy-efficiency, and area utilization of our design on two different types of Field-Programmable Gate Array (FPGAs) as well as using the OpenROAD and mapped against two different standard cell libraries: FreePDK-45nm and ASAP 7nm, and we
\item  Finally, we demonstrate two proof-of-concept applications running using our IzhiRISC-V system, showing how our extensions can be used to improve performance on a cortical microcircuit simulation as well as solving a constraint satisfaction problem (CSP) such as Sudoku.
\end{itemize}

The paper is structured as follows: Section \ref{sec:background} provides information about neuromorphic computing, Spiking Neural Networks (SNNs) and implementing software and hardware SNN simulators, Section \ref{sec:rel_works} presents similar works towards neuromorphic ISA extensions for RISC-V processors, Section \ref{sec:isa} presents the proposed neuromorphic ISA extension for Izhikevich neurons, Section \ref{sec:arch} showcases the implemented IzhiRISC-V processor supporting the aforementioned ISA extension, \ref{sec:results} presents the implementation details in hardware of the IzhiRISC-V-based system and presents test scenarios, and Section \ref{sec:conclusion} concludes the paper. 
\section{Background}
\label{sec:background}
\subsection{Neuromorphic computing}
\label{sec:background_neuro}
Neuromorphic systems, introduced for the first time in the 1980s by Carver Mead\cite{mead2002neuromorphic}, are computers that utilize models of computation present in the biological brain. Initially designed to replicate brain circuitry using analog electronic components, neuromorphic systems today are implemented in different technologies, ranging from digitally synchronous (e.g., Intel Loihi~\cite{davies2018loihi} or IBM TrueNorth~\cite{akopyan2015truenorth}), mixed analog/digital (e.g., BrainDrop~\cite{neckar2018braindrop}), fully analog (e.g. \cite{indiveri2011neuromorphic}), to even memristor- or photonics-based systems~\cite{li2018review}. Those systems are different from classic von-Neumann-like computers in a sense that they perform computations using neurons and synapses, communicate with spikes and are programmed through encoding the problem to be solved by a Spiking Neural Network (SNN) instead of relying on a central processing unit executing imperative instructions stored in memory, and they are implemented as bespoke hardware for such processing\cite{szczerek2025quarter}. 

In the digital synchronous domain, the two most important representatives of neuromorphic processing units are Intel Loihi and its successor - Loihi 2\footnote{https://www.intel.com/content/www/us/en/research/neuromorphic-computing.html}) - as well as IBM TrueNorth.  The IBM TrueNorth is the first truly scalable digital neuromorphic chip supporting up to one million LIF neurons with 256 million synapses. Multiple chips can be strapped together to expand the processing capabilities. The newer version of the Loihi chip allows for simulating networks of up to one million neurons and 120 million synapses through 128 neural cores, as well as supporting programmable pipelines for neuron models, i.e., it is not bound to a single model like Leaky Integrate-and-Fire (LIF), as was the case with its predecessor. It is also scalable, allowing for connecting multiple chips together to expand the SNN processing capabilities. There is also yet another approach to the topic in the form of SpiNNaker\cite{furber2012overview} and SpiNNaker 2\cite{mayr2019spinnaker}, which merge the general-purpose processing capabilities of an array of 153 ARM M4F cores with dedicated per-core hardware pipelines for SNN computing.

\subsection{Spiking Neural Networks}
\label{sec:background_snn}
Biological neurons communicate via \textit{action potentials}, commonly referred to as \textit{spikes}. Networks of neurons that model this behavior are called \textit{Spiking Neural Networks} (SNNs), often called the \textit{third generation networks}\cite{maass1997networks}. SNNs have been used as tools to study the biological brain through simulators such as Neuron\cite{hines1997neuron}, NEST\cite{diesmann2001nest}, or Brian\cite{stimberg2019brian}, as well as for solving machine learning (ML) problems (classification tasks\cite{wang2022triplebrain, carpegna2024spiker+, li2021fast}, robotics\cite{mitchell2017neon, gomez2016ed}). Crucially, SNNs are believed to reduce the energy consumption needed to solve those problems and/or solve them faster, primarily by leveraging the sparseness of spike communication.

SNNs are composed of spiking neurons, which pass the information between each other via spikes traveling through synapses. The biological properties of synapses are often modeled by setting a specific \textit{synaptic weight} to a connection, which, in connection with the spike traveling through it, results in a change of synaptic current at the postsynaptic side of the connection. This change influences the state of the neuron - primarily the membrane potential - and makes the postsynaptic neuron more or less likely to spike\cite{szczerek2025quarter}.

There are many different spiking neuron models in common use today, varying in terms of biological plausibility, complexity, and, by implication, intended use cases. Those models range from simple Leaky Integrate-and-Fire (LIF), which describes neuronal dynamics with a rather simple ordinary differential equation (ODE) and is commonly used in Computer Science, to the complex Hodgkin-Huxley (HH) model that aims at representing a plethora of different neuronal behaviors by describing said dynamics with a set of four ODEs. Those two extremes of the spiking neuron models spectrum are of great importance, and there are multiple well-known systems that use them. However, they do suffer from being "extreme" - LIF generally cannot represent more complicated behaviours on a single neuron level, and HH, due to its complexity, is sparsely used outside of the neuronal simulators, built specifically to observe complex interactions between neurons. The Izhikevich model (IZH) can be viewed as a middle ground between those two extremes - it is able to represent a number of neuronal behaviours, while requiring rather low processing, at the cost of biological plausibility - i.e., the actual biological parameters are not \textit{directly} representable in this model. There are two main variants of the IZH - simpler with four and complex with nine parameters - and the former takes the form of a set of two ODEs of membrane potential variable $v$ and recovery variable $u$. When the membrane potential is below the \textit{threshold voltage} $V_{th}$, the equation is as follows:
\begin{equation}
\label{eq:izh}
\begin{cases}
   \frac{dv}{dt} = 0.04v^2 + 5v + 140 - u + I_{syn} \\
   \frac{du}{dt} = a(bv - u)
\end{cases}
\end{equation}
and when membrane potential reaches the threshold $(v > V_{TH})$, the variables are set to the following values:
\begin{equation}
    \begin{cases}
        v \leftarrow c \\
        u \leftarrow u + d
    \end{cases}
\end{equation}
where $a$, $b$, $c$, and $d$ are specific parameters which allow for obtaining different neuronal spiking behaviors (Tonic Spiking, Intrinsic Bursting, and so on) and $I_{syn}$ is the presynaptic current. Moreover, the $c$ variable can be understood as the reset voltage $V_{RST}$ that the actual biological neurons achieve after they spike. The neuron is also generating a \textit{spike} at this point, which is being received by every neuron connected to the spiking neuron. This spike causes a change in the synaptic current, which influences the $v$ and $u$ variables of the postsynaptic neurons.

\subsection{Digital SNNs in software and hardware}
The most obvious way of implementing SNNs in a digital manner is creating a software program targeting a general-purpose CPU/GPU system that iterates over all neurons and updates their state by a predetermined timestep $h$. However, if we consider a general-purpose CPU, such as RISC-V, and a \textit{Forward Euler} approximation scheme (which is commonly used in literature\cite{nanami2016fpga, farsa2019low}), then for every neuron update, we need to perform the following calculations
\begin{equation}
\label{eq:izh_num}
    \begin{cases}
   v_{n+1} = (0.04v_n^2 + 5v_n + 140 - u_n + I_{syn})h + v_n \\
   u_{n+1} = ah(bv_n - u_n) + v_n
\end{cases}
\end{equation}
which results in 15 operations for every neuron. In the equation above $h$ is the predetermined timestep over which we want to update the neuronal state, $n$ denotes the current and $n+1$ the next global timestep, i.e. the moment in time of the simulation. As neuron update is the costliest \textit{basic} operation in SNN simulation and has to be performed at every timestep progression, we can deduce that using standard CPU instructions (like \textbf{add}, \textbf{mul} or \textbf{sub}) may be inefficient due to instruction repetition. Moreover, it is common in SNN to model the synaptic current decay, which is an actual biological phenomenon\cite{purves_neuroscience_2008}, which can be modeled as an exponential decay
\begin{equation}
\label{eq:dec}
    \frac{I_{syn}}{dt} = -\frac{I_{syn}}{\tau}
\end{equation}
where $\tau$ is the decay constant, related to, among other features of the neuronal system, the resistance and capacitance of the neuronal membrane\cite{purves_neuroscience_2008}. After applying the \textit{Forward Euler} approximation method, we obtain the following update equation
\begin{equation}
\label{eq:dec_num}
    I_{syn_{n+1}} = -\frac{I_{syn_n}}{\tau}h
\end{equation}
with $h$ denoting the predetermined timestep. This results in an additional four operations required for every synaptic current decay, and in summary, 19 operations are required to perform the necessary neuron-related operations. Packaging those operations and performing them as one operation should noticeable performance improvements in terms of SNN simulations on RISC-V processors.

On the other hand, one can implement a bespoke hardware processing element (PE) that would perform the necessary computations after a request sent from a CPU, in a System-on-a-Chip (SoC) manner. This approach has the potential of reaching significant improvement in performance in comparison to the previous method - especially if the entire neuron update pipeline is implemented via non-standard, optimized hardware. The PEs used in this method typically leverage the on-chip resources to further speed up the computations\cite{szczerek2025quarter}. Two examples of such an approach were listed in Section \ref{sec:background_neuro} - Intel Loihi 1/2 and IBM TrueNorth.

There is also a method that stands in the middle of the two approaches, i.e., by converting the general-purpose processor into an Application Specific Instruction set Processor (ASIP) via custom instructions that deal with the non-standard operations, such as IZH neuron update. By doing so, the natural programming flow of the CPU is unobstructed and does not rely on the external PEs, and the complexity and size of the CPU core would not be increased by a significant amount. An example of such a system using such a principle would be the aforementioned SpiNNaker 1/2 systems. In this work, we followed this design principle, which we will describe in Sections \ref{sec:isa} and \ref{sec:arch}.
\section{Related works}
\label{sec:rel_works}
Neuromorphic extensions for RISC-V ISA have been getting traction in the research community in recent years, as there are various approaches to the topic presented in the literature. In this Section, we list the most prominent examples and underline their characteristics.

Zhang et al. \cite{zhang2025asynchronous} presented a neuromorphic extension to a five-stage-pipeline RISC-V processor based on the so-called "clock-joint" circuits - a specific communication scheme allowing for asynchronous operation. Their implementation assumed utilizing a bespoke SNN core, capable of implementing Feed-Forward topology with 512 LIF neurons in the hidden layer and 16 in the output layer, or 256 neurons connected in a recurrent manner. The core implemented a bespoke hardware pipeline for neurons, spike input packet processing, and memory for the network parameters with 8-bit arithmetic. 
Wang et al. \cite{wang2024rv} presented an ISA extension with multi-cycle instructions targeting primarily Spiking Convolutional Neural Networks (SCNNs) and regular CNNs. The design assumed utilizing a custom co-processor with support for instructions tailored for use in CNN-like networks, e.g., converting input images to column vectors. The co-processor consisted of a 4x4 grid of Processing Elements (PEs) that calculated the necessary LIF neuronal dynamics in SNN mode. Liu et al. \cite{liu2024activen} presented a scalable RISC-V system with support for non-ad-hoc hardware extensions to allow the SNN through an active-message-enabled architecture to support the event-driven programming model.
Mendat et al. \cite{mendat2023risc} presented a mixed-signal RISC-V-based system, based on a SiFive RISC-V core, and integrating an analog Compute-in-Memory core for SNN computations. This implementation allowed for analog-domain computations of the SNN-related dynamics.
Yang et al. \cite{yang2023back} presented a NeuroRV processor, based on an open-source RI5CY\cite{gautschi2017near} core. They introduced a vector extension to the ISA for performing synaptic current accumulation, neuron state updates, MAC operations, and classification. Their NeuroRV core utilized additional cores for vector instructions and processing a number of neurons simultaneously. The implementation presented in the paper allowed for 1024 simultaneous neuron updates through a matching number of processing elements.
Forno et al. \cite{forno2021configuring} presented a system utilizing a RISC-V-compliant Rocket Chip processor \cite{asanovic2016rocket} and ODIN neuromorphic co-processor\cite{frenkel20180}, where both chips communicated via an SPI bus, making it the most SoC-like implementation out of the ones listed in this Section. The implementation allowed for simultaneous update of up to 256 LIF or modified IZH neurons, as compliant with ODIN's specifications.
Manoni et al. \cite{manoni2025spikestream} presented a SpikeStream concept, aiming at accelerating SNN-based inference on clusters of RISC-V processors via an ISA extension for streaming sparse computations. The authors claimed to have achieved  4.39× speedup and an increase in effective resource utilization from 9.28\% to 52.3\% compared to a "non-streaming parallel baseline" for the VGG-11 model.
\section{ISA Extension for Izhikevich neurons}
\label{sec:isa}

To enhance the capability of RISC-V-based systems in terms of processing SNNs, we propose a custom neuromorphic instruction set. Those instructions are RISC-compliant, i.e., they can be executed in a single clock cycle and refer to the basic operations required to update spiking neurons by a single timestep. The four proposed instructions use the \textit{custom-0} opcode, i.e., "0001011", as stated in the official RISC-V ISA and are listed in Table \ref{tab:isa}. The first pair is referred to as \textbf{configuration instructions}, while the latter are the \textbf{processing instructions}.

The system is designed with signed 16-bit fixed-point arithmetic in mind, which is reflected in the input/output data format of the instructions. All of the parameters and values used in SNN processing are of 16-bit precision, with a specific Q-format selected for different operands to maximize the accuracy of the results. The specific Q-formats for the operands are listed in Table \ref{tab:isa}. 

\begin{table}[htbp]
\caption{Custom ISA extension - opcode: 0001011}
\begin{center}
\begin{tabular}{|c||c|c|c|}
\hline
\textbf{funct3} & \textbf{rs2} & \textbf{rs1} &  \textbf{rd} \\
\textbf{[14:12]} & \textbf{[24:20]} & \textbf{[19:15]} &  \textbf{[11:7]} \\
\hline
\hline
&$31...16: d$ & $31...16: b$ & \\
& $(Q4.11)$ & $(Q4.11)$ & \\
$nmldl$ & $15...0: c$ & $15...0: a$ &  $dst: 1 = OK$ \\
$(R)$& $(Q7.8)$ & $(Q4.11)$ & \\
\hline
& &$31...2: resv.$ & \\
$nmldh$ & $reserved$ & $1: pin$ & $dst: 1 = OK$\\
$(R)$& & $0: h$ & \\
\hline
 &  & $31...16: v$ & $src:addr(VU)$ \\
$nmpn$ & $I_{syn}$ & $(Q7.8)$ & $dst: spike$ \\
$("N")$ & $(Q15.16)$ & $15...0: u$ & $(1 - spike$ \\
&  & $(Q7.8)$ &  $0 - no \ spike)$ \\
\hline
$nmdec$ & $\tau \ select$ &$I_{syn}$ & $dst: dec(I_{syn})$ \\
$(R)$ & $(1...9)$ & $(Q15.16)$ & $(Q15.16)$ \\
\hline
\end{tabular}
\begin{tabular}{|c||c|c|c|}
\multicolumn{1}{l}{$^{\mathrm{1}}$If $pin$ bit is set, NPU caps the neuron voltage at reset potential.}\\
\multicolumn{1}{l}{$^{\mathrm{2}}$If $h$ bit is set, the NPU operates with timestep of 0.125ms else - 0.5ms.}\\
\multicolumn{1}{l}{$^{\mathrm{3}}$The $nmpn$ instruction performs Izhikevich neuron update for $v$ and $u$}\\
\multicolumn{1}{l}{variables with timestep corresponding to $h$. The updated neuron state is }\\ 
\multicolumn{1}{l}{stored back under the address passed through rd register. The rd register}\\
\multicolumn{1}{l}{holds the result -- information on the neuron spiking in the current}\\
\multicolumn{1}{l}{timestep.}\\
\multicolumn{1}{l}{$^{\mathrm{4}}$The $nmdec$ instruction performs exponential decay operation on Q15.16}\\ 
\multicolumn{1}{l}{value with timestep corresponding to $h$: $y(x) = -(1/2^\tau)xh$.}
\end{tabular}
\label{tab:isa}
\end{center}
\end{table}

\subsection{Configuration instructions}
The configuration instructions are used to set the necessary parameters of the Izhikevich neuron being processed by the NPU. They follow the \textit{R-type} format, where source registers rs1 and rs2 hold the necessary parameters, and the result is a simple flag that indicates the information is stored in the NPU's registers. The \textbf{nmldl} instruction sets the $a$, $b$, $c$, $d$ parameters of the Izhikevich neuron, which were described in Section \ref{sec:background}. The parameters use the Q4.11 representation for $a$, $b$, and $d$ parameters, while the $c$ parameter is in Q7.8 format. The \textbf{nmldh} instruction sets the required biological timestep - the user can choose between 0.5 ms and 0.125 ms - and whether to prevent the membrane voltage from reaching values lower than reset voltage $V_{RST}$ or not through setting the $pin$ bit.

\begin{lstlisting}[float, floatplacement=hbtp,language={[RISC-V]Assembler}, caption=RISC-V assembly code utilizing the custom instructions.]
    lw a6, 4(a3)
    lw a7, 8(a3)
    nmldl x0, a6, a7 # load a,b,c,d parameters
    lw t5, (a4)	# read the thalamic
    lw a7, (a0)	# read current
    lw a6, (a3) # read vu
    add a7, a7, t5
    add a2, x0, a3
    nmpn a2, a6, a7 #process neuron, get spike/nospike, store VU word
\end{lstlisting}

\subsection{Processing instructions}
The processing instructions are used to advance the state of the Izhikevich neuron by the predetermined timestep with preset neuron parameters (set by the configuration instructions). The \textbf{nmpn} instruction follows a non-standard \textit{"N"-type}, which allows for simultaneously storing the result in memory, while returning $1$ if a spike occurred in the current timestep or $0$ if it did not. The instruction accepts three pieces of information, due to treating the rd register as a source register in the fetch/decode stages and as a destination register in the execution stage and onward, namely a \textit{VU word}, which holds Izhikevich neuron variables $v$ and $u$ in Q7.8 format in a single 32-bit word, synaptic current $I_{syn}$ in rs2 in Q15.16 format and 32-bit address of the \textit{VU word}. The \textbf{nmdec} instruction provides exponential decay capability with a pre-configured timestep used for synaptic current decay
\begin{equation}
    dI_{syn} = -\frac{I_{syn}}{2^{-\tau}}h
\end{equation}
where $I_{syn}$ is the synaptic current, $\tau$ is configurable in the interval $[1,9]$. 
\section{Architecture}
\label{sec:arch}
In this Section, we present the \textbf{IzhiRISC-V} core, based on the base RISC-V core called \textit{DTEK-V}.
\begin{figure}[htbp]
\centerline{\includegraphics[width=0.5\textwidth]{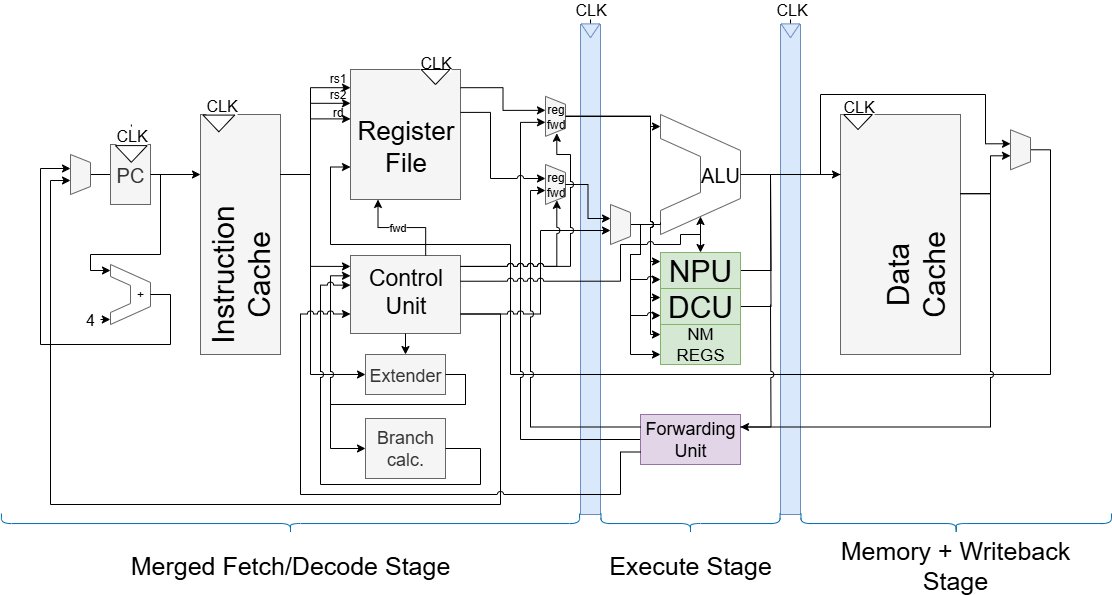}}
\caption{Overall architecture of the IzhiRISC-V core, with added NPU and DCU marked in green.}
\label{fig:dtekv}
\end{figure}

\subsection{DTEK-V}
The basic DTEK-V core is compatible with RV32IMZ ISA, and its architecture is presented in Figure \ref{fig:dtekv} (excluding the blocks coloured in green). Instead of aiming for a standard 5-stage pipeline design, a few improvements were introduced, most prominently \textit{merging the Fetch and Decode stages} and \textit{merging the Memory and Writeback stages}, thus making it a 3-stage pipeline. The architecture also includes a Forwarding Unit, which allows for forwarding data from the Execute and Memory + Writeback stages to the Register File. The Forwarding Unit also includes circuitry to prevent data hazards via inserting delay cycles into pipeline operation.

\subsection{Neuromorphic extensions units - NPU and DCU}
\label{sec:arch_npudcu}
To allow the aforementioned ISA extension to function, we implemented a small Neuron Processing Unit (NPU) and Neuron Decay Unit (DCU), which both are extensions of the DTEK-V's ALU, instead of making up a separate co-processor, as portrayed in Figure \ref{fig:dtekv} - with NPU and DCU marked in green. That being said, they allow for single-cycle operation not to interfere with the natural programming flow of a RISC-like processor, providing simple instructions for basic operations for processing SNNs - neuronal update and synaptic decay. The NPU and DCU support two timestep values - 0.5ms and 0.125ms, which allowed us to use bit-shift operations instead of costly divisions - and rely on the information stored in the configuration registers, visible in Figure \ref{fig:dtekv} as \textit{NM REGS}. Those registers need to be preloaded with the necessary IZH parameters - $a$, $b$, $c$, $d$, $h$, $\tau$ and the \textit{pin voltage bit} as defined in Sections \ref{sec:background_snn} and \ref{sec:isa}, prior to NPU and/or DCU can operate on a requested neuron. That is achieved through using \textbf{nmldl} and \textbf{nmldh} instructions. When the \textbf{nmpn} instruction is called, the NPU performs the IZH neuron update as described in equation \ref{eq:izh_num} and returns two pieces of information - updated IZH variables as the \textit{VU word} and sets a bit in another word if the spike occurred. The operation of this instruction is somewhat unorthodox because it works like a merge of an R-type and an S-type instruction. As it was listed in \ref{tab:isa}, apart from the \textit{VU word} and synaptic current $I_{syn}$, the address of the \textit{VU word} is also passed. NPU performs computations on the two former operands, while ALU computes the destination address for the Memory + Writeback stage to store the result in memory. It is worth mentioning that we are considering using CSR registers for writeback of the spike information from \textbf{nmpn} and the \textit{done} flags from the \textbf{nmldl} and \textbf{nmldh} instructions instead of following the classic flow of writing this information to the register file. That way, we can potentially significantly speed up the simulation as the information about spikes is usually volatile and used primarily to modify the synaptic current of the postsynaptic neurons. However, at this stage of the project, we were more concerned with the correctness of the results rather than performance optimizations.

Both NPU and DCU were designed directly in RTL using VHDL and a fixed-point package provided in the IEEE library - we represented the values as type $sfixed$ with appropriate bit sizes of integer and mantissa for every parameter and value used in computations.

The calculations performed in the NPU for $v$ and $u$ variables are done with a variable size of the accumulator, because different operands use different fixed-point formats - listed in Table \ref{tab:isa}, and the resulting values of the variables are converted to the default format of Q7.8. If the \textit{pin voltage bit} is set, the NPU additionally will set $v \leftarrow c$ if the resulting change in $v$ variable would make it lower than $V_{RST}$. This behavior prevents the rebound property of the original IZH model\cite{izhikevich2004model}, and we added it to the NPU, as we observed a potential decrease in solution convergence for one of the use cases - the Sudoku solver. We decided not to use any optimizations towards resource utilization in $v$ and $u$ update computations, such as replacing multiplications with additions and bit-shifts, as we focused on ensuring the correctness of the results rather than speed of operation at this stage of the project. We managed to achieve single-cycle operation of the NPU with a clock frequency of 30MHz - same as the clock frequency of the original DTEK-V processor implemented on the Intel MAX10 board.

The computations performed in the DCU are less complicated, as can be seen in the Equation \ref{eq:dec_num}. Instead of actual division circuitry, DCU implements a division approximator, which uses bit shift and addition operators. The following operations are only possible because we are utilizing fixed-point representation, which stores the values as integers that are later interpreted to obtain the fixed-point values. The input value is shifted by a factor from one to nine and then, based on the $\tau$ specified in the \textbf{nmdec} instruction, appropriate bit-shifted values are summed up to obtain the approximated division. For example, suppose we want to calculate $\frac{x}{7}$, which is approximately equal to $0.1428571x$. The closest value to $0.1428571$ that can be obtained using the bit-shift array with factors from one to nine is $(x\gg3 + x\gg6 + x\gg9)=0.125x + 0.015625x + 0.0019353125x = 0.142578125x$, which results in an approximation error (AE) of
\begin{equation}
    AE = \frac{0.142578125 - \frac{1}{7}}{\frac{1}{7}}*100\% \approx 0.1953\%
\end{equation}
The DCU supports division approximation for dividers in the interval from $/2$ to $/8$, and the values of $AE$ are visible in Table \ref{tab:dcu_approx_error}. Apart from the $/6$ division is reaching $AE$ values lower than $0.5\%$, which we tested to be satisfactory for the SNN simulation.

\begin{table}[htbp]
\caption{Approximation errors for the DCU for dividers from $/2$ to $/8$ with available bit shifters with factors from one to nine.}
\begin{center}
\begin{tabular}{|c||c|c|}
\hline
\textbf{Division} & \textbf{Approx. value} & \textbf{AE [$\approx\%$]} \\
\hline
$x/2$ & $x \gg 1$ & 0 \\
\hline
$x/3$ & $(x \gg 2) + (x \gg 4) + (x \gg 6) + (x \gg 8)$  & $0.3906$ \\
\hline
$x/4$ & $x \gg 2$ & 0 \\
\hline
$x/5$ & $(x \gg 3) + (x \gg 4) + (x \gg 7) + (x \gg 8)$  & $0.3906$ \\
\hline
$x/6$ & $(x \gg 3) + (x \gg 5) + (x \gg 7) + (x \gg 9)$  & $12.1093$ \\
\hline
$x/7$ & $(x \gg 3) + (x \gg 6) + (x \gg 9)$ & $0.1953$ \\
\hline
$x/8$ & $x \gg 3$ & 0 \\
\hline
\end{tabular}
\label{tab:dcu_approx_error}
\end{center}
\end{table}

\section{Results and Discussion}
\label{sec:results}
The architecture described in Section \ref{sec:arch} was tested on a low-end Intel MAX10 FPGA with two use cases: an "80-20" SNN simulation (80\% excitatory, 20\% inhibitory neurons) used in neuroscience to observe population spiking behaviors of groups of neurons, and solving Top 100 difficult Sudoku problems. As we are presenting the first steps towards a larger system, the performance was not our priority - we focused on the correctness of the implementation and estimating the potential scalability of the future system. In both presented use cases, we were able to fit the network-related data on the on-chip memory, while the instructions were fetched from the off-chip SDRAM. 

\subsection{FPGA-based implementation}
The implementation was tested on the TerasIC DE10-Lite board with Intel MAX10 FPGA (10M50DAF484C7G), where we were comfortably able to fit 2 IzhiRISC-V cores clocked at 30MHz, connected to a common Avalon bus. We utilized on-chip memory for necessary state variables to reduce costly access to external SDRAM memory. The dual-core system occupied virtually all of the available logic resources on the MAX10 device, as can be seen in Table \ref{tab:max10_util}. It is worth noting that we were able to put three cores on the said device, but we needed to drastically reduce the size of both data and instruction cache and reduce the clock frequency to 20 MHz, which had a detrimental impact on performance.

\begin{table}[htbp]
\caption{Resource utilization for dual-core IzhiRISC-V system on Intel MAX10 device.}
\begin{center}
\begin{tabular}{|c||c|}
\hline
Frequency & 30MHz \\
\hline
Logic elements & 49248 (99\%) \\
\hline
FF  & 28235 (51\%) \\
\hline
BRAM & 346.468Kb (21\%) \\
\hline
Embedded Mult. (9b) & 68 (24\%) \\
\hline
\end{tabular}
\label{tab:max10_util}
\end{center}
\end{table}

Moreover, we synthesized 3 IzhiRISC-V systems with 10, 32, and 64 cores for Intel Agilex-7 FPGA M-Series Development Kit with Intel Agilex-7 FPGA (AGMF039R47A1E2VR0) to get an estimate on the scale of a hypothetical many-core IzhiRISC-V system on an HPC-grade FPGA. The results in Table \ref{tab:agilex7_util} suggest that we can create a system with up to 192 cores on this kind of platform running at 100MHz, assuming linear scaling. However, we acknowledge the fact that with this number of cores, the interconnect, data exchange between cores, and memory access are crucial factors, and to fully benefit from the performance of such a system, a different type of connectivity is in order, e.g., using a Network-on-a-Chip (NoC) structure in place of a common bus.

\begin{table}[htbp]
\caption{Resource utilization for 16-, 32- and 64-core IzhiRISC-V system on Intel Agilex-7 device (based on Intel's Golden Hardware Reference Design (GHRD)).}
\begin{center}
\begin{tabular}{|c||c|c|c|}
\hline
\textbf{Metric} & \textbf{16 cores} & \textbf{32 cores} &  \textbf{64 cores} \\
\hline
Frequency & \multicolumn{3}{c|}{100 MHz} \\
\hline
ALM & 107144 (8\%) & 216448 (17\%) & 420977 (32\%) \\
\hline
FF  & 95624 (2\%) & 186760 (4\%) & 372741 (7\%) \\
\hline
RAM blocks & 390 (2\%) & 646 (3\%) & 1158 (6\%) \\
\hline
DSP & 152 (1\%) & 304 (2\%) & 608 (5\%) \\
\hline
\end{tabular}
\label{tab:agilex7_util}
\end{center}
\end{table}

\subsection{Izhikevich's 80-20 network}
\begin{figure}[htbp]
\centerline{\includegraphics[width=0.5\textwidth]{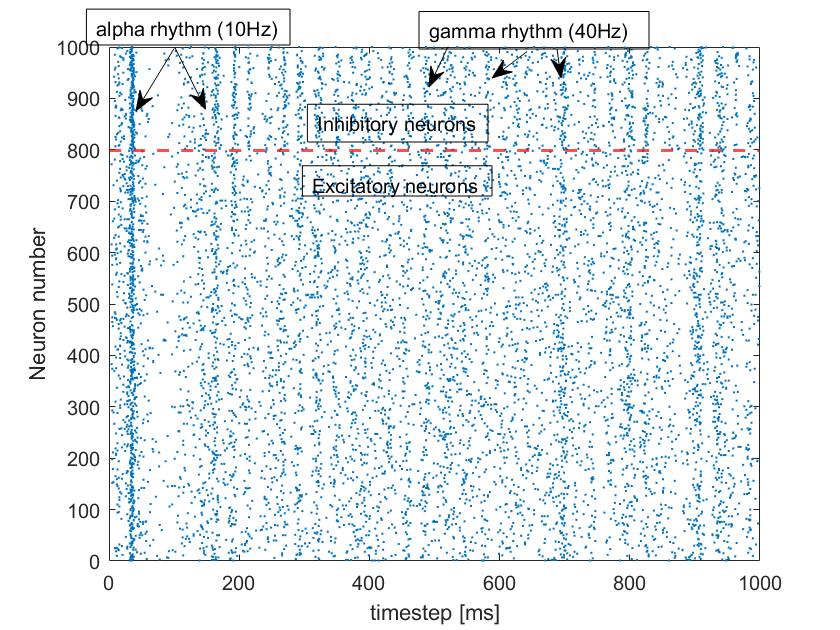}}
\caption{Raster plot of the 80-20 IZH network simulation.}
\label{fig:raster}
\end{figure}

The first use case we selected to test the implementation was simulating 1000 IZH neurons ($80\%$ excitatory and $20\%$ inhibitory) with random connections with timestep $h = 1ms$. This network was derived from Izhikevich's work\cite{izhikevich2003simple}, where the original double-precision-based setup was converted to an appropriate 16-bit fixed-point representation (following the formats presented in Table \ref{tab:isa}) and moved to the IzhiRISC-V system. The simulated network exhibited similar alpha/gamma rhythms, visible in the raster plot of spiking neurons in Figure \ref{fig:raster}, as the MATLAB implementations\footnote{The actual raster plots for the double-precision implementation can be found in the original work\cite{izhikevich2003simple}.} (double precision and fixed-point representation, similar to that in IzhiRISC-V). Moreover, the histogram of the inter-spike-intervals (ISI) for those three implementations suggests similar simulation results, i.e., similar operation in terms of generating the alpha and gamma rhythms in the neuron population. The dual-core implementation achieved a speedup of 1.643x over a single-core IzhiRISC-V setup, achieving an effective instruction-per-cycle (IPC) of around 0.66, where:
\begin{equation}
    IPC = \frac{N_{instr}}{N_{cycles}}
\end{equation}
where $N_{instr}$ is the number of instructions executed and $N_{cycles}$ is the number of clock cycles that the CPU took to execute $N_{instr}$ instructions. Ideal $IPC$ is equal to 1. The effective IPC $IPC_{eff}$ is calculated as follows:
\begin{equation}
    IPC_{eff} = \frac{N_{reginstr} + N_{updates}N_{IZHop}}{N_{cycles}}
\end{equation}
where $N_{reginstr}$ is the number of regular (not-related to actual IZH neuron update) instructions, $N_{updates}$ is the number of neuronal updates, $N_{IZHop}$ is the number of necessary equivalent regular instructions to perform the neuronal update (i.e. update of the $v$ and $u$ variables by timestep $h$, as well as the synaptic current decay) and $N_{cycles}$ is the number of clock cycles (all $N_x$ values are gathered over the main loops of the programs when the neurons are being updated). As it was explained in Section \ref{sec:background_snn}, $N_{IZHop} = 19$. The $IPC_{eff}$ can be larger than 1.

\begin{table}[htbp]
\caption{Performance metrics for 80-20 network (1000 neurons, 1000 timesteps, 1ms timestep) for one and two cores clocked at 30MHz.}
\begin{center}
\begin{tabular}{|c||c|c|c|}
\hline
& &  \multicolumn{2}{c|}{\textbf{Dual-core}}\\\cline{3-4}
\textbf{Metric} & \textbf{Single-core} & \textbf{Core \#1} & \textbf{Core \#2}\\
\hline
Speedup & 1 & \multicolumn{2}{c|}{1.643x}\\
\hline
Execution time [s] & 7.870 & 4.791 & 4.7906\\
\hline
$IPC$ & 0.5735 & 0.5317 & 0.51887\\
\hline
$IPC_{eff}$ & 0.6516 & 0.6637 & 0.6508\\
\hline
Hazard stalls [\%] & 0.742 & 5.344 & 6.259\\
\hline
All cache misses & 1306420 & 639798 & 675623\\
\hline
I-cache hit rate [\%] & 99.97 & 99.97 & 99.97 \\
\hline
D-cache hit rate [\%] & 96.54 & 97.18 & 97.09 \\
\hline
Mem intensity & 27.15 & 28.88 & 30.12 \\
\hline
\end{tabular}
\label{tab:80_20_perf}
\end{center}
\end{table}

The gathered metrics suggest that the implementation running on the dual-core system is not operating at its full potential, as $IPC \approx 0.5$ and $IPC_{eff} < 1$. We expect it is due to the present hazard stalls, resulting when the pipeline is halted if any of the source registers of the fetched instruction is equal to the destination register of the current instruction, which can be remedied by the usage of \textit{CSR writeback} for the proposed instructions as described in Section \ref{sec:arch_npudcu}. Moreover, the memory layout and parameter storage can be improved to reduce the overall cache miss rate and lower the access times. However, as we stated previously, performance optimizations were not our main goal in the work towards this project so far.

\begin{figure}[htbp]
\centering 
\includegraphics[width = 0.22\textwidth]{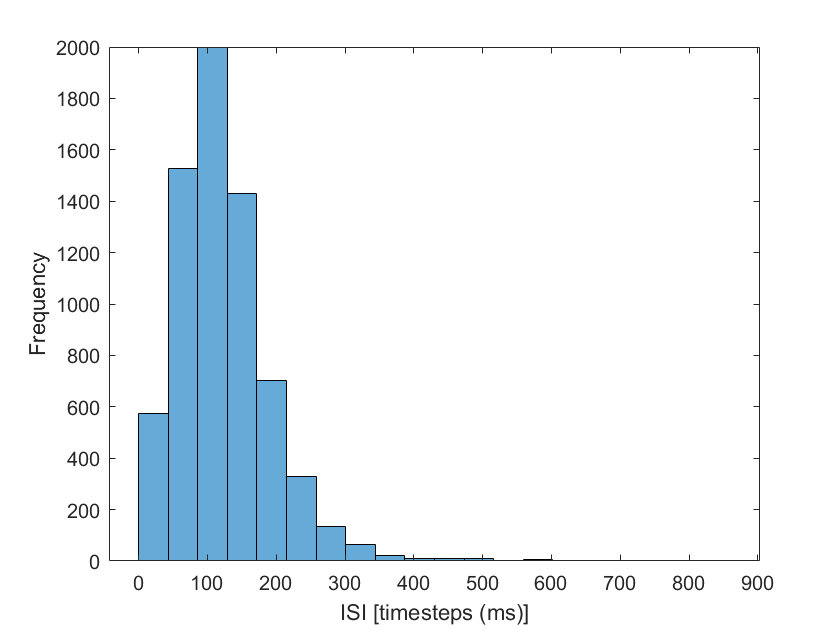}\quad
\includegraphics[width = 0.22\textwidth]{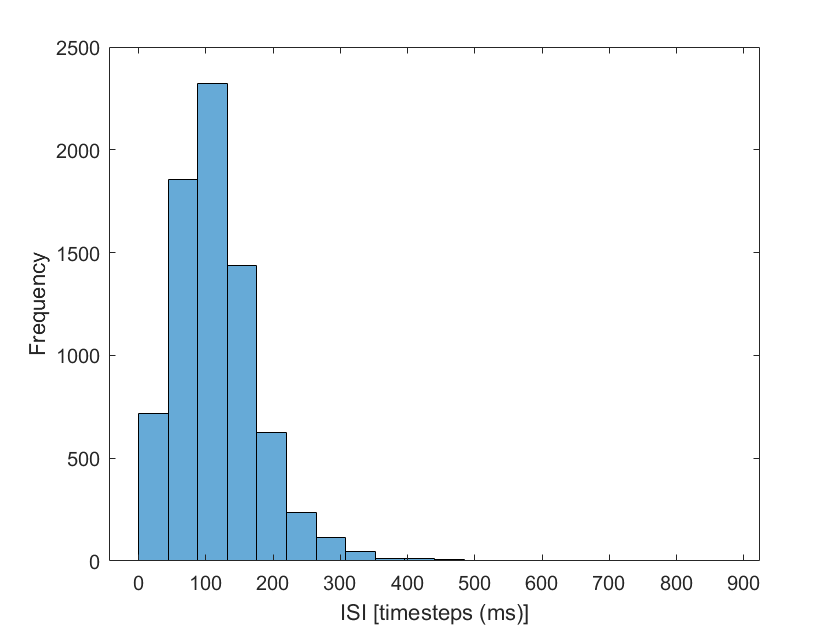}\par\medskip
\includegraphics[width = 0.22\textwidth]{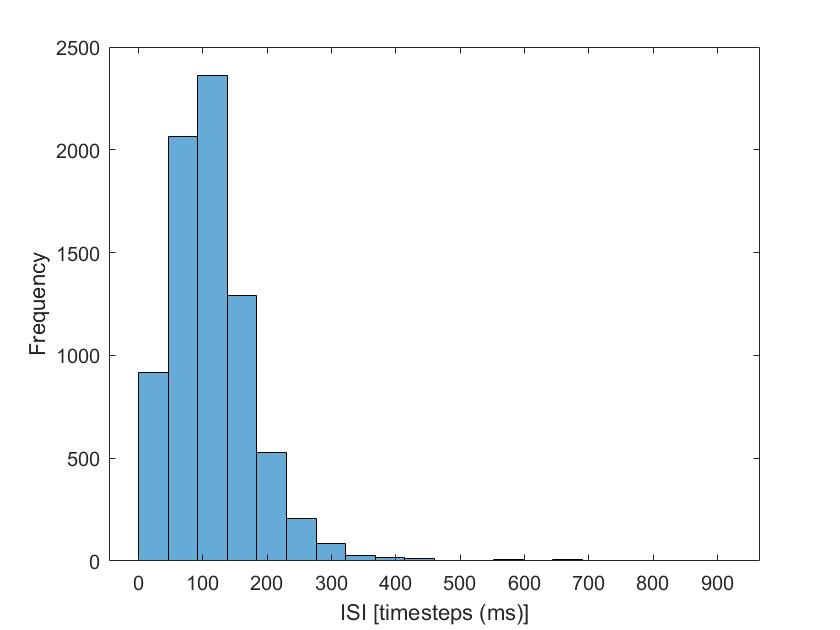}
\caption{Comparison of ISI histograms for different implementations of 80-20 network (top row - MATLAB (left - double precision, right - fixed-point), bottom row - IzhiRISC-V.}
\end{figure}

\subsection{Sudoku solver}
\begin{figure}[htbp]
\centerline{\includegraphics[width=0.5\textwidth]{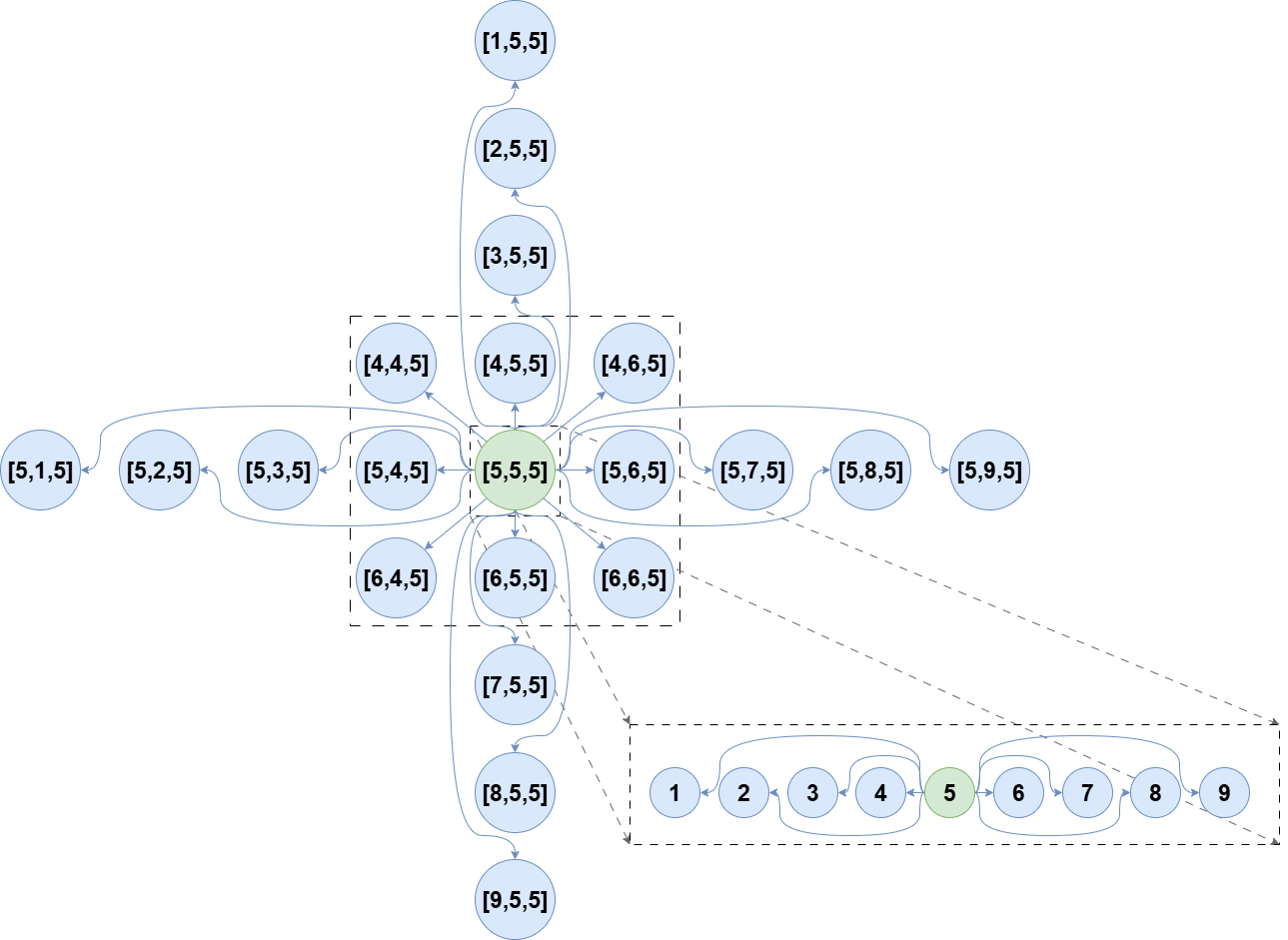}}
\caption{Graph showing the inhibitory connections from an example spiking neuron (green) to other neurons in the "multi-level" WTA network in Sudoku solver (denoted in blue). The dashed lines around the 3x3 grid relate to the 3x3 subgrid of the 9x9 Sudoku board. The indices [x,y,z] indicate the row and column of a particular cell within the 9x9 Sudoku board, and the last index indicates which digit (1-9) the neuron is supposed to represent. The small subnetwork on the bottom right relates to the digits 1-9.}
\label{fig:sudoku_network}
\end{figure}

The Sudoku solver use case relied on a network of 729 IZH neurons, arranged in a particular example of a Winner-Takes-All (WTA) network. Every cell in a 9x9 Sudoku grid "consists" of an array of nine neurons, every one representing a digit from one to nine. If a particular neuron from this array spikes, it inhibits the neurons responsible for the same digit in the same row, column, and small 3x3 subgrid the spiking neuron belongs to - visualization of this inhibition process is visible in Figure \ref{fig:sudoku_network}.

We tested this network with Top 100 difficult Sudoku puzzles\footnote{http://magictour.free.fr/top100}, all of which were solved by the system. The used NPU and DCU utilizing the fixed-point arithmetic described in Section \ref{sec:isa} allowed for significant improvements in computation speed in comparison to soft-float implementation supported by original DTEK-V, without sacrificing the required accuracy of neural simulation to solve the Sudoku puzzles - we observed an approx. 40x reduction in execution time per timestep. We do realize that comparing the performance against the soft-float implementation is not ideal, as IzhiRISC-V does not have a floating-point unit and thus any computations using single-precision floating-point numbers, especially the division, can be very slow. However, our fixed-point Sudoku solver without NPU and DCU was not able to always converge at the correct solution, primarily due to the necessary conversion of the synaptic current from 32-bit fixed-point to 16-bit fixed-point value, which, in this implementation, did not round the value correctly. We acknowledge that an improved fixed-point solver implementation is necessary to fully benchmark the performance of the solver.

The gathered metrics suggest, similar to the 80-20 use case, that the cores are not utilized to their full potential, as $IPC_{eff} < 1$. We can also see the hazard stalls being present, which we will try to resolve with \textit{CSR writeback}. Moreover, the memory organization will always be rethought for the larger system.

\begin{table}[htbp]
\caption{Performance metrics for the Sudoku solver (729 neurons, 1ms timestep) for one and two cores clocked at 30MHz. As different Sudoku puzzles required various numbers of timesteps, an average value (marked with $AVG$) over all puzzles or an average per-timestep (marked with $TSTP$) value is provided where applicable.}
\begin{center}
\begin{tabular}{|c||c|c|c|}
\hline
& &  \multicolumn{2}{c|}{\textbf{Dual-core}}\\\cline{3-4}
\textbf{Metric} & \textbf{Single-core} & \textbf{Core \#1} & \textbf{Core \#2}\\
\hline
Speedup & 1 & \multicolumn{2}{c|}{1.682x}\\
\hline
Execution time [ms] $(TSTP)$ & 2.0555 & 1.2223 & 1.2223 \\
\hline
$IPC$ $(AVG)$ & 0.5304 & 0.4960 & 0.4194\\
\hline
$IPC_{eff}$ $(AVG)$ & 0.7564 & 0.8635 & 0.7865\\
\hline
Hazard stalls [\%] $(AVG)$ & 5.136 & 6.4793 & 9.1493\\
\hline
I-cache hit rate [\%] $(AVG)$  & 98.7230 & 98.6848 & 98.8331 \\
\hline
D-cache hit rate [\%] $(AVG)$ & 99.9999 & 100.0 & 99.9999 \\
\hline
Mem intensity $(AVG)$ & 21.3853 & 22.3176 & 23.9244 \\
\hline
\end{tabular}
\label{tab:sudoku}
\end{center}
\end{table}

\subsection{Mapping to standard cell library}
In order to explore the hypothetical performance of our design if taped out, we map an IzhiRISC-V processing core to the 45 nm FreePDK45\footnote{https://eda.ncsu.edu/freepdk/freepdk45/} and ASAP7 standard cell library using the open source OpenROAD~\cite{ajayi2019openroad} design flow, using GHDL\footnote{http://ghdl.free.fr/} as a preprocessing step to convert VHDL to Verilog.

\begin{figure}[ht]
\begin{minipage}[b]{0.45\linewidth}
\centering
\includegraphics[width=\textwidth]{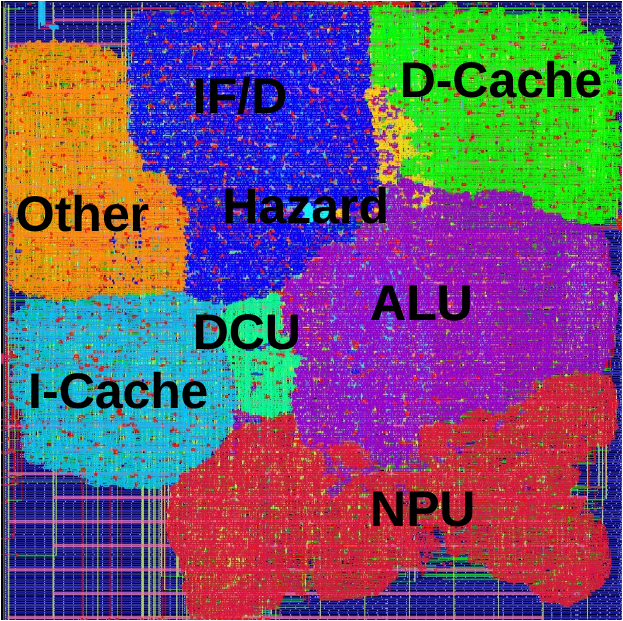}
\end{minipage}
\hspace{0.5cm}
\begin{minipage}[b]{0.45\linewidth}
\centering
\includegraphics[width=\textwidth]{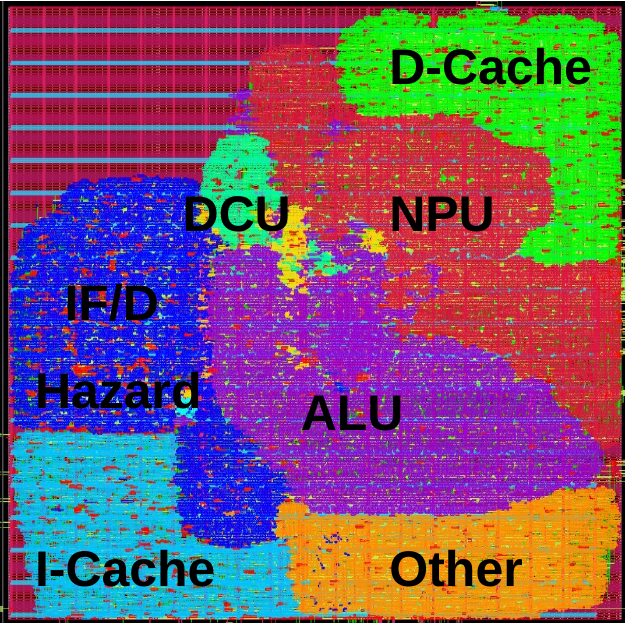}
\end{minipage}
\caption{Final floorplan of an IzhiRISC-V core on both the FreePDK 45 nm (left) and ASAP 7 nm (right) process, showing the NPU and DCU in comparison to other blocks in the pipeline.}
\label{fig:stdcell}
\end{figure}

\begin{table}[htbp]
\caption{FreePDK45 and ASAP7 Standard Cell Mapping Results}
\begin{center}
\begin{tabular}{|c || l | c | c | l |}
\hline
\textbf{} & \textbf{Metric} & \textbf{FreePDK7} & \textbf{ASAP7} & \textbf{unit}\\
\hline

\multirow{9}{*}{\rotatebox{90}{\textbf{Area}}} 
& Total area                & $95654.664$  & 6599.375 & $\mu m^2$\\ \cline{2-5}
& Fetch/Decode              & $16924.250$  & 1116.522 & $\mu m^2$\\ \cline{2-5}
& Instruction Cache         & $10588.662$  & 723.941  & $\mu m^2$\\ \cline{2-5}
& Data Cache                & $12097.414$  & 799.830  & $\mu m^2$\\ \cline{2-5}
& Hazard Unit               & $146.300 $   & 7.480    & $\mu m^2$\\ \cline{2-5}
& ALU                       & $19873.924 $ & 1441.364 & $\mu m^2$\\ \cline{2-5}
& NPU                       & $19516.154 $ & 1292.196 & $\mu m^2$\\ \cline{2-5}
& DCU                       & $2005.640$   & 141.411  & $\mu m^2$\\ \cline{2-5}
& Other                     & $11449.172$  & 809.584  & $\mu m^2$\\ \hline
\hline
\multirow{4}{*}{\rotatebox{90}{\textbf{Power}}} 
& Total Power               & $49.5$           & $10.9$ & $mW$ \\ \cline{2-5}
& Internal                  & $25.7 (51.9\%)$  & $6.05 (55.5\%)$ & $mW$ \\ \cline{2-5}
& Switching                 & $21.5 (43.5\%)$  & $4.85 (44.5\%)$  & $mW$ \\ \cline{2-5}
& Leakage                   & $2.31 (4.7\%)$   & $ 6.45 (0.1\%)$ & $\mu W$ \\ \hline
\hline

\multirow{3}{*}{\rotatebox{90}{\textbf{Speed}}} 
& Clock freq.               &    $201.5$     & $316.3$ & $MHz$\\ \cline{2-5}
& Throughput                &    $67.6$      & $105.4$ & $MUpd./s$\\ \cline{2-5}
& Power efficiency:         &    $1.371$     & $9.67$  & $GUpd/s/W$\\ \cline{2-5}
& Peak neural IPS\footnote{This performance is for repeatedly updating the same neuron}:          &    $3.022$     & $4.74$ & $GInstr./s$ \\ \hline

\end{tabular}
\label{tab:stdcell}
\end{center}
\end{table}

The place-and-routed version of the core using the both libraries is seen in Figure~\ref{fig:stdcell}, with the different components inside the IzhiRISC-V processing core highlighted. Overall, our NPU additions occupy no more that roughly 20\% of the entire core area, while our DCU occupies even less ($<2\%)$, and would likely occupies even less with an increase data- or instruction-cache size. The estimated power consumption for the core is far less than $100 mW$, allowing for high theoretical power-efficient performance (up to $4.74$ giga-neuron updates per second and watt). Our IzhiRISC-V processing core can be clocked at 316.3 MHz (7 nm) and would reach a theoretical peak performance of 105 million neural updates per second.
\section{Conclusion}
\label{sec:conclusion}
We proposed an ISA extension for Izhikevich neuron computation for RISC-V general-purpose processor, as well as showcased an implementation of the IzhiRISC-V processor on low-end Intel MAX10 processor and tested it with two neuromorphic-related use cases - 80-20 network simulation and Sudoku solving through an SNN. We estimated the possibility of synthesizing a large system with 192 cores on Intel Agilex-7 and also mapped the dual-core IzhiRISC-V system to FreePDK45 45 nm process and ASAP7 standard library using OpenROAD. We gathered important insight into continuing the design of a large IzhiRISC-V-based system on HPC-grade FPGAs.

 \bibliographystyle{ieeetr}
 \bibliography{ref}

\begin{thebibliography}{10}

\bibitem{bohr200930}
M.~Bohr, ``A 30 year retrospective on dennard's mosfet scaling paper,'' {\em IEEE Solid-State Circuits Society Newsletter}, vol.~12, no.~1, pp.~11--13, 2009.

\bibitem{theis2017end}
T.~N. Theis and H.-S.~P. Wong, ``The end of moore's law: A new beginning for information technology,'' {\em Computing in science \& engineering}, vol.~19, no.~2, pp.~41--50, 2017.

\bibitem{shalf2020future}
J.~Shalf, ``The future of computing beyond moore’s law,'' {\em Philosophical Transactions of the Royal Society A}, vol.~378, no.~2166, p.~20190061, 2020.

\bibitem{schuman2017survey}
C.~D. Schuman, T.~E. Potok, R.~M. Patton, J.~D. Birdwell, M.~E. Dean, G.~S. Rose, and J.~S. Plank, ``A survey of neuromorphic computing and neural networks in hardware,'' {\em arXiv preprint arXiv:1705.06963}, 2017.

\bibitem{szczerek2025quarter}
W.~J. Szczerek and A.~Podobas, ``A quarter of a century of neuromorphic architectures on fpgas--an overview,'' {\em arXiv preprint arXiv:2502.20415}, 2025.

\bibitem{maass1997networks}
W.~Maass, ``Networks of spiking neurons: the third generation of neural network models,'' {\em Neural networks}, vol.~10, no.~9, pp.~1659--1671, 1997.

\bibitem{davies2021advancing}
M.~Davies, A.~Wild, G.~Orchard, Y.~Sandamirskaya, G.~A.~F. Guerra, P.~Joshi, P.~Plank, and S.~R. Risbud, ``Advancing neuromorphic computing with loihi: A survey of results and outlook,'' {\em Proceedings of the IEEE}, vol.~109, no.~5, pp.~911--934, 2021.

\bibitem{izhikevich2003simple}
E.~M. Izhikevich, ``Simple model of spiking neurons,'' {\em IEEE Transactions on neural networks}, vol.~14, no.~6, pp.~1569--1572, 2003.

\bibitem{mead2002neuromorphic}
C.~Mead, ``Neuromorphic electronic systems,'' {\em Proceedings of the IEEE}, vol.~78, no.~10, pp.~1629--1636, 2002.

\bibitem{davies2018loihi}
M.~Davies, N.~Srinivasa, T.-H. Lin, G.~Chinya, Y.~Cao, S.~H. Choday, G.~Dimou, P.~Joshi, N.~Imam, S.~Jain, {\em et~al.}, ``Loihi: A neuromorphic manycore processor with on-chip learning,'' {\em Ieee Micro}, vol.~38, no.~1, pp.~82--99, 2018.

\bibitem{akopyan2015truenorth}
F.~Akopyan, J.~Sawada, A.~Cassidy, R.~Alvarez-Icaza, J.~Arthur, P.~Merolla, N.~Imam, Y.~Nakamura, P.~Datta, G.-J. Nam, {\em et~al.}, ``Truenorth: Design and tool flow of a 65 mw 1 million neuron programmable neurosynaptic chip,'' {\em IEEE transactions on computer-aided design of integrated circuits and systems}, vol.~34, no.~10, pp.~1537--1557, 2015.

\bibitem{neckar2018braindrop}
A.~Neckar, S.~Fok, B.~V. Benjamin, T.~C. Stewart, N.~N. Oza, A.~R. Voelker, C.~Eliasmith, R.~Manohar, and K.~Boahen, ``Braindrop: A mixed-signal neuromorphic architecture with a dynamical systems-based programming model,'' {\em Proceedings of the IEEE}, vol.~107, no.~1, pp.~144--164, 2018.

\bibitem{indiveri2011neuromorphic}
G.~Indiveri, B.~Linares-Barranco, T.~J. Hamilton, A.~v. Schaik, R.~Etienne-Cummings, T.~Delbruck, S.-C. Liu, P.~Dudek, P.~H{\"a}fliger, S.~Renaud, {\em et~al.}, ``Neuromorphic silicon neuron circuits,'' {\em Frontiers in neuroscience}, vol.~5, p.~73, 2011.

\bibitem{li2018review}
Y.~Li, Z.~Wang, R.~Midya, Q.~Xia, and J.~J. Yang, ``Review of memristor devices in neuromorphic computing: materials sciences and device challenges,'' {\em Journal of Physics D: Applied Physics}, vol.~51, no.~50, p.~503002, 2018.

\bibitem{furber2012overview}
S.~B. Furber, D.~R. Lester, L.~A. Plana, J.~D. Garside, E.~Painkras, S.~Temple, and A.~D. Brown, ``Overview of the spinnaker system architecture,'' {\em IEEE transactions on computers}, vol.~62, no.~12, pp.~2454--2467, 2012.

\bibitem{mayr2019spinnaker}
C.~Mayr, S.~Hoeppner, and S.~Furber, ``Spinnaker 2: A 10 million core processor system for brain simulation and machine learning-keynote presentation,'' in {\em Communicating Process Architectures 2017 \& 2018}, pp.~277--280, IOS Press, 2019.

\bibitem{hines1997neuron}
M.~L. Hines and N.~T. Carnevale, ``The neuron simulation environment,'' {\em Neural computation}, vol.~9, no.~6, pp.~1179--1209, 1997.

\bibitem{diesmann2001nest}
M.~Diesmann and M.-O. Gewaltig, ``Nest: An environment for neural systems simulations,'' {\em Forschung und wisschenschaftliches Rechnen, Beitr{\"a}ge zum Heinz-Billing-Preis}, vol.~58, pp.~43--70, 2001.

\bibitem{stimberg2019brian}
M.~Stimberg, R.~Brette, and D.~F. Goodman, ``Brian 2, an intuitive and efficient neural simulator,'' {\em elife}, vol.~8, p.~e47314, 2019.

\bibitem{wang2022triplebrain}
H.~Wang, Z.~He, T.~Wang, J.~He, X.~Zhou, Y.~Wang, L.~Liu, N.~Wu, M.~Tian, and C.~Shi, ``Triplebrain: A compact neuromorphic hardware core with fast on-chip self-organizing and reinforcement spike-timing dependent plasticity,'' {\em IEEE Transactions on Biomedical Circuits and Systems}, vol.~16, no.~4, pp.~636--650, 2022.

\bibitem{carpegna2024spiker+}
A.~Carpegna, A.~Savino, and S.~Di~Carlo, ``Spiker+: a framework for the generation of efficient spiking neural networks fpga accelerators for inference at the edge,'' {\em IEEE Transactions on Emerging Topics in Computing}, 2024.

\bibitem{li2021fast}
S.~Li, Z.~Zhang, R.~Mao, J.~Xiao, L.~Chang, and J.~Zhou, ``A fast and energy-efficient snn processor with adaptive clock/event-driven computation scheme and online learning,'' {\em IEEE Transactions on Circuits and Systems I: Regular Papers}, vol.~68, no.~4, pp.~1543--1552, 2021.

\bibitem{mitchell2017neon}
J.~P. Mitchell, G.~Bruer, M.~E. Dean, J.~S. Plank, G.~S. Rose, and C.~D. Schuman, ``Neon: Neuromorphic control for autonomous robotic navigation,'' in {\em 2017 IEEE International Symposium on Robotics and Intelligent Sensors (IRIS)}, pp.~136--142, IEEE, 2017.

\bibitem{gomez2016ed}
F.~G{\'o}mez-Rodr{\'\i}guez, A.~Jim{\'e}nez-Fernandez, F.~P{\'e}rez-Pe{\~n}a, L.~Mir{\'o}, M.~J. Dom{\'\i}nguez-Morales, A.~Rios-Navarro, E.~Cerezuela, D.~Cascado-Caballero, and A.~Linares-Barranco, ``Ed-scorbot: A robotic test-bed framework for fpga-based neuromorphic systems,'' in {\em 2016 6th IEEE international conference on biomedical robotics and biomechatronics (BioRob)}, pp.~237--242, IEEE, 2016.

\bibitem{nanami2016fpga}
T.~Nanami and T.~Kohno, ``An fpga-based cortical and thalamic silicon neuronal network,'' {\em Journal of Robotics, Networking and Artificial Life}, vol.~2, no.~4, pp.~238--242, 2016.

\bibitem{farsa2019low}
E.~Z. Farsa, A.~Ahmadi, M.~A. Maleki, M.~Gholami, and H.~N. Rad, ``A low-cost high-speed neuromorphic hardware based on spiking neural network,'' {\em IEEE Transactions on Circuits and Systems II: Express Briefs}, vol.~66, no.~9, pp.~1582--1586, 2019.

\bibitem{purves_neuroscience_2008}
D.~Purves, G.~J. Augustine, D.~Fitzpatrick, W.~C. Hall, A.-S. LaMantia, J.~O. McNamara, and L.~E. White, eds., {\em Neuroscience, 4th Ed.}
\newblock Neuroscience, 4th Ed., Sunderland, MA, US: Sinauer Associates, 2008.

\bibitem{zhang2025asynchronous}
X.~Zhang, J.~Zhang, H.~Huang, and H.~Chen, ``An asynchronous risc-v-based snn processor with custom isa extensions for programmable on-chip learning,'' in {\em 2025 29th IEEE International Symposium on Asynchronous Circuits and Systems (ASYNC)}, pp.~1--8, IEEE, 2025.

\bibitem{wang2024rv}
X.~Wang, C.~Feng, X.~Kang, Q.~Wang, Y.~Huang, and T.~T. Ye, ``Rv-scnn: A risc-v processor with customized instruction set for snn and cnn inference acceleration on edge platforms,'' {\em IEEE Transactions on Computer-Aided Design of Integrated Circuits and Systems}, 2024.

\bibitem{liu2024activen}
X.~Liu, Z.~Pu, P.~Qu, W.~Zheng, and Y.~Zhang, ``Activen: A scalable and flexibly-programmable event-driven neuromorphic processor,'' in {\em 2024 57th IEEE/ACM International Symposium on Microarchitecture (MICRO)}, pp.~1122--1137, IEEE, 2024.

\bibitem{mendat2023risc}
D.~R. Mendat, J.~P. Sengupta, G.~Tognetti, M.~Villemur, P.~O. Pouliquen, S.~Montano, K.~Sanni, J.~L. Molin, N.~Zachariah, I.~Doxas, {\em et~al.}, ``A risc-v neuromorphic micro-controller unit (vmcu) with event-based physical interface and computational memory for low-latency machine perception and intelligence at the edge,'' in {\em 2023 IEEE International Symposium on Circuits and Systems (ISCAS)}, pp.~1--5, IEEE, 2023.

\bibitem{yang2023back}
Z.~Yang, L.~Wang, W.~Shi, Y.~Wang, J.~Tie, F.~Wang, X.~Yu, L.~Peng, C.~Xiao, X.~Xiao, {\em et~al.}, ``Back to homogeneous computing: A tightly-coupled neuromorphic processor with neuromorphic isa,'' {\em IEEE Transactions on Parallel and Distributed Systems}, vol.~34, no.~11, pp.~2910--2927, 2023.

\bibitem{gautschi2017near}
M.~Gautschi, P.~D. Schiavone, A.~Traber, I.~Loi, A.~Pullini, D.~Rossi, E.~Flamand, F.~K. G{\"u}rkaynak, and L.~Benini, ``Near-threshold risc-v core with dsp extensions for scalable iot endpoint devices,'' {\em IEEE transactions on very large scale integration (VLSI) systems}, vol.~25, no.~10, pp.~2700--2713, 2017.

\bibitem{forno2021configuring}
E.~Forno, A.~Spitale, E.~Macii, and G.~Urgese, ``Configuring an embedded neuromorphic coprocessor using a risc-v chip for enabling edge computing applications,'' in {\em 2021 IEEE 14th International Symposium on Embedded Multicore/Many-core Systems-on-Chip (MCSoC)}, pp.~328--332, IEEE, 2021.

\bibitem{asanovic2016rocket}
K.~Asanovic, R.~Avizienis, J.~Bachrach, S.~Beamer, D.~Biancolin, C.~Celio, H.~Cook, D.~Dabbelt, J.~Hauser, A.~Izraelevitz, {\em et~al.}, ``The rocket chip generator,'' {\em EECS Department, University of California, Berkeley, Tech. Rep. UCB/EECS-2016-17}, vol.~4, pp.~6--2, 2016.

\bibitem{frenkel20180}
C.~Frenkel, M.~Lefebvre, J.-D. Legat, and D.~Bol, ``A 0.086-mm2 12.7-pj/sop 64k-synapse 256-neuron online-learning digital spiking neuromorphic processor in 28-nm cmos,'' {\em IEEE transactions on biomedical circuits and systems}, vol.~13, no.~1, pp.~145--158, 2018.

\bibitem{manoni2025spikestream}
S.~Manoni, P.~Scheffler, L.~Zanatta, A.~Acquaviva, L.~Benini, and A.~Bartolini, ``Spikestream: Accelerating spiking neural network inference on risc-v clusters with sparse computation extensions,'' in {\em 2025 Design, Automation \& Test in Europe Conference (DATE)}, pp.~1--7, IEEE, 2025.

\bibitem{izhikevich2004model}
E.~M. Izhikevich, ``Which model to use for cortical spiking neurons?,'' {\em IEEE transactions on neural networks}, vol.~15, no.~5, pp.~1063--1070, 2004.

\bibitem{ajayi2019openroad}
T.~Ajayi and D.~Blaauw, ``Openroad: Toward a self-driving, open-source digital layout implementation tool chain,'' in {\em Proceedings of Government Microcircuit Applications and Critical Technology Conference}, 2019.

\end{thebibliography}

\end{document}